\documentclass[runningheads,a4paper]{llncs}

\usepackage{amssymb}
\usepackage{graphicx}
\usepackage{subfig}
\usepackage{url}
\usepackage{verbatim}
\usepackage{hyperref}

\usepackage{amsmath}

\newcommand{\B}{\emph{forward }}
\newcommand{\A}{\emph{backward }}
\newcommand{\hbi}{$h_F(s_i)$}

\newcommand{\haj}{$h_B(s_j)$}
\newcommand{\si}{s_i}
\newcommand{\sj}{s_j}
\newcommand{\argmax}{\operatornamewithlimits{argmax}}
\newcommand{\biB}{\mathbf b_{iF}}
\newcommand{\bjA}{\mathbf b_{jB}}
\newcommand{\BB}{\mathbf B_{F}}
\newcommand{\BA}{\mathbf B_{B}}
\newcommand{\WB}{\mathbf W_F}
\newcommand{\WA}{\mathbf W_B}

\begin{document}

\mainmatter  

\title{Deep Segment Hash Learning for Music Generation}

\titlerunning{Deep Segment Hash Learning for Music Generation}

%
%

\author{Kevin Joslyn, Naifan Zhuang \and Kien A. Hua}
%
\authorrunning{Kevin Joslyn, Naifan Zhuang and Kien A. Hua}

	
\institute{University of Central Florida \\ \email{KevinJoslyn@knights.ucf.edu} \\
\email{zhuangnaifan@knights.ucf.edu} \\
\email{kienhua@cs.ucf.edu}}

	
%
%

\maketitle

\begin{abstract}


Music generation research has grown in popularity over the past decade, thanks to the deep learning revolution that has redefined the landscape of artificial intelligence. In this paper, we propose a  novel approach to music generation inspired by musical segment concatenation methods and hash learning algorithms. Given a segment of music, we use a deep recurrent neural network and ranking-based hash learning to assign a \emph{forward} hash code to the segment to retrieve candidate segments for continuation with matching \emph{backward} hash codes. The proposed method is thus called Deep Segment Hash Learning (DSHL). To the best of our knowledge, DSHL is the first end-to-end segment hash learning method for music generation, and the first to use pair-wise training with segments of music. We demonstrate that this method is capable of generating music which is both original and enjoyable, and that DSHL offers a promising new direction for music generation research.
\end{abstract}

\section{Introduction}

Music is often called the ``universal language" because it is something that every human being can understand and appreciate. While each individual person may have different musical tastes, we generally agree upon what sounds good and what does not. Music composers harness this innate sense of musical essence along with their own creativity and experience in order to create new and interesting songs that the audience can enjoy and appreciate. Generally, the underlying rules governing how composers create this music must be learned through years of studying music theory, listening to music, and writing original music. Teaching a computer how to simulate musical creativity is thus no easy task, and it is one of the great challenges to artificial intelligence.

The idea of computer-generated music is intriguing for a variety of reasons. First, there is a virtually infinite amount of unique music that can be created, so computer-generated music can serve to expand the scope of all imaginable music and create original pieces that reach all audiences with varying musical tastes and cultures. Second, music generation software can serve to complement composers and actually aid them in their own compositions. Indeed, this was the motivation in \cite{cope} when the author set out to create a program that could help free him from his own writer's block. The resulting Experiments in Musical Intelligence system \cite{cope} was at the time the most thorough attempt to generate music using a computer, and he accomplished it without the use of the deep learning resources that are so widely available today. The original approach to music generation in \cite{cope} took a database of many songs and split the songs up into many small parts, or \emph{segments}, with the goal of recombining the segments in an original way to create new music. The resulting piece retained the style of the music in the database, but was unrecognizable because it purposefully limited the repeated use of segments from the same song. However, there are a few drawbacks to the method, including the use of many ad-hoc rules and segment modifications, as well as the need for manually labeled segments. In this paper we introduce a new model that does not require any user-defined rules or manually labeled segments. Instead, we allow a neural network to implicitly learn the rules of music theory by training it on many examples from a dataset of music conforming to a given style.

Since the pioneering work in \cite{cope}, computer-generated music has become a very popular research area. Segment concatenation methods remain of interest because all segments already come from valid existing music, and each segment conforms to the style of music that is desired. Existing works \cite{unitsel,improtek,continuator,hashingcont} all use this approach; however, these methods often do not have efficient means to store and retrieve good candidate segments from the database. \cite{hashingcont} solves this problem by assigning hash codes to segments of music, and using these hash codes to query the database for continuation segments. However, the hash functions are not directly learned from the data. Instead, the hashing method assumes that the database is already stored in a tree structure, where child nodes represent possible continuations of the parent sub-tree. In \cite{continuator,hashingcont}, such a tree is constructed by linearly traversing the dataset and directly placing segments onto the tree, after applying a reduction function to the segment to allow searching for content that is similar to the query. The main drawback of this approach is that the creativity of system is limited to the generalizability of the reduction function, since the retrieved continuations must have been directly observed in the dataset.


Motivated by the idea of fast retrieval of continuations using discrete learned hash codes, we desire a segment hashing approach that leverages deep learning to improve the creativity of the generated music. Rather than restricting the possible continuations to those that are observed in the dataset, our goal is to teach a neural network by example to understand the underlying rules that govern good segment-to-segment transitions. Thus, we propose the first end-to-end hash learning solution for musical segment concatenation, called Deep Segment Hash Learning (DSHL). The high-level objective of DSHL is to learn a hash function that hashes similar segments to the same hash bucket, so that we can efficiently retrieve possible continuations of a query segment by using a corresponding query hash code.




To determine which segments are suitable candidates to follow a given segment of music, we need a hash function that can capture the \emph{composability} of a pair of segments (i.e. the likelihood that a segment could follow another given segment in a piece of music). This requirement leads us to design a parallel neural network that supports pair-wise training, given a pair of segments and a \emph{composability} label. To the best of our knowledge this type of pair-wise training has not yet been applied to music generation despite being commonly used in hashing for similarity search methods \cite{lsrh,dcmh,ddsrh}. During training, these related hashing methods take an input pair of database items (generally consisting of an image and/or text,) and either minimize or maximize the difference between the computed hash codes depending on whether the pair of items are related. For our method, we can naively say that a training pair of segments is a \emph{composable} pair if the second segment is seen occurring after the first segment in any song in the database, and not composable otherwise. Later, we will discuss exceptions to this rule as it relates to the training of our network.

The DSHL hash function learning is accomplished by utilizing recurrent neural networks (RNN)  \cite{rnn}, which have frequently been used in music generation research due to their ability to model time-series data and predict the next timestep in a sequence. Specifically, Long Short-Term Memory networks (LSTM)  \cite{lstm}, a variant of RNN's, are more often used due to their ability to remember long histories of events and capture global structure in music \cite{lstm02,tiedparallel,pi,jambot,crnngan}. DSHL uses LSTM's to discover an intermediate representation of the segments that is used by the remainder of the hashing network.


In summary, the major contributions of the paper are as follows:

\begin{itemize}
	\item We propose a novel approach to music generation via segment concatenation that is the first end-to-end hash learning method for musical segment retrieval.
	\item We introduce the first music generation system that learns hash functions via pair-wise training using ``composable" and ``non-composable" pairs.
	\item We demonstrate that the generated music sounds pleasing to the ear, and that the proposed hashing method assigns similar hash codes to composable pairs and dissimilar hash codes to non-composable pairs.
\end{itemize}

The remainder of the paper is structured as follows. In section \ref{sec:relatedwork}, we cover related work, focusing on methods that use segment concatenation, methods that use RNN's, as well as related hashing methods for multimedia retrieval. In section \ref{sec:method}, we discuss DSHL in detail, including the neural network architecture and the loss function that governs the hash function learning. In section \ref{sec:experiments}, we discuss the experiments and results, and in section \ref{sec:conclusions} we conclude the paper and discuss future work.







\section{Related Work} \label{sec:relatedwork}


Our method is most closely related to a class of music generation algorithms called segment concatenation methods. These methods compose music by concatenating ``building blocks" of short musical segments that are stored in a database of segments. Primarily, segment concatenation methods must determine a solution for how to select a suitable ``next" segment. Another important issue arises in how to efficiently search the database for the continuation segment.

The EMI project \cite{cope} labels each segment as a ``statement", ``preparation", ``extension", ``antecedent", or ``consequent," and uses an augmented transition network (ATN) to develop an automated grammar to compose music via segment concatenation. \cite{unitsel} defines both a ``semantic relevance cost" and a ``concatenation cost" which are used to evaluate candidate segments. The semantic relevance cost is calculated using a deep structured semantic model \cite{dssm}, while the concatenation cost is calculated by using a note-level LSTM that predicts the next note in the melody. Since the method does not use smart indexing of the segments, it must search through the entire database of segments in order to rank the candidates, which can be costly. ImproteK \cite{improtek} searches its database to find a segment that shares both a common history and a common future with the current segment. It uses a prefix indexing technique to avoid unnecessary comparisons through brute force linear search through the memory. The Continuator \cite{continuator} is a segment concatenation method that uses a pattern-discovery Markov model to construct a tree-structure database where child nodes represent possible continuations for a parent segment. However, storing all possible continuations for a segment in a tree is prohibitive for large databases. \cite{hashingcont} proposes a solution to this problem by computing hash codes for the edges in the tree, as well as hash codes for paths that are calculated by appending several edge hashes. The result is a reduced database size (since common subtrees hash to the same value), and a fast lookup of possible continuations. Our method differs greatly from \cite{hashingcont} and the aforementioned techniques because the hashing algorithm is not built on top of an existing model of segment composability. Rather, the hash function is directly learned in order to preserve the pair-wise composability of two segments, while ensuring fast and efficient retrieval of the next segment.


The origin of hashing for similarity search was Locality Sensitive Hashing \cite{lsh}, which introduced the idea of purposefully using hash collisions to store similar database instances in the same hash bucket. Our proposed hashing method is actually based upon Deep De-correlated Subspace Ranking Hashing (DDSRH) \cite{ddsrh}, a deep cross-modal hashing method originally designed to generate hash codes that preserve the cross-modal similarity between images and texts. DDSRH, like its predecessor Linear Subspace Ranking Hashing (LSRH) \cite{lsrh}, generates a $K$-ary hash code by first learning a unique $K$-dimensional subspace for each hash ``bit". Each hash bit is then determined by projecting the original features onto the corresponding subspace and taking the index of the dimension with the maximum projection. DDSRH benefits from the fact that the subspaces are nonlinear due to the use of deep neural networks, and each hash bit is de-correlated due to the lack of interconnections between the neurons that determine each subspace. Since the idea of DDSRH is not specific to images and texts, the hashing method is easily applied to musical segments analyzed by an LSTM, CNN, etc.


Many methods have used RNN's and RNN variants (especially the LSTM) to generate music by means of timestep prediction \cite{lstm02,boulanger,rnnrbm,tiedparallel,algornn,waite,brnn,concert,pi,jambot,crnngan}. Similar to our method, DeepBach \cite{deepbach} uses both a forward RNN and a backward RNN. However, these RNN's are used to predict a pitch given the history and future of the surrounding musical context respectively. Our work is the first to combine the sequence modeling capabilities of RNN's with hash learning for music generation.






\section{Deep Segment Hash Learning} \label{sec:method}

In this section we describe Deep Segment Hash Learning (DSHL) in detail. In section \ref{sec:method1}, we discuss the \B and \A hash code scheme, and give a brief overview of the neural network that learns the hash functions. In section \ref{sec:lstm}, we discuss how the model uses LSTM's for sequence modeling. In section \ref{sec:rankhash}, we describe the ranking-based hashing scheme that we use in this work and how the architecture of the network allows it to learn the hash functions. In section \ref{sec:objfunc}, we discuss the objective function which we desire to minimize during training. Finally, in section \ref{sec:training}, we explain in detail how the network is trained using pairs of segments.

\subsection{Forward and Backward Hash Codes} \label{sec:method1}



Similar to other segment concatenation methods \cite{cope,unitsel,improtek,continuator,hashingcont}, the main challenge facing DSHL is how to determine suitable candidate segments to follow the most recent segment in the generation process. Taking inspiration from \cite{hashingcont}, we are motivated to solve this problem by assigning hash codes to each segment, so that continuation segments can be retrieved by using the query hash code of the most recent segment. In this way, candidate segments can be retrieved using an efficient K-Nearest Neighbors (KNN) search on data with discrete values.


Unlike \cite{hashingcont}, which assigned hash codes to paths in a tree of continuations observed in the dataset, we desire a more generalizable model that leverages deep learning to understand what comprises a good segment-to-segment transition. To accomplish this, the model must be able to accurately grade the \emph{composability} of any pair of segments. Given a pair of segments ($\si$,$\sj$), we say that ($\si$,$\sj$) is a \emph{composable} pair if $\sj$ is suitable to follow $\si$ in a piece of music. Now, in order to design a hash function that facilitates query and retrieval of segments based on their composability, we must consider the difference between the query segment $\si$ and a possible retrieval segment $\sj$. The query segment should be analyzed in the forward direction (forward in time), since the music is heading towards a following idea or arrival point. On the other hand, the retrieval segment should reflect in some way on the preceding query segment, i.e. it should logically follow the previous idea. Thus we can analyze the retrieval segment by taking the retrograde (reverse in time) of the segment, in order to essentially predict what came before.




Following this intuition, we propose to give each segment in the database a \B and a \A hash code, each obtained by analyzing the segment in the forward and reverse direction in time, respectively. Given a segment $s$, we denote the \B hash code as $h_F(s)$ and the \A hash code as $h_B(s)$. If a segment pair ($\si$,$\sj$) is a \emph{composable} segment pair, we would like the hamming distance between $h_F(s_i)$ and $h_B(s_j)$ to be minimized. Otherwise, if it is a non-composable pair, the hamming distance should be maximized. In essence, we can think of these hash codes as being representations for a certain ``virtual state" that exists between a pair of segments. If $h_F(s_i) \approx h_B(s_j)$, we have found a common representation for the virtual state between $\si$ and $\sj$; thus, the $\si \rightarrow \sj$ transition is admissible.

\begin{figure}
	\centering
	\includegraphics[height=6cm]{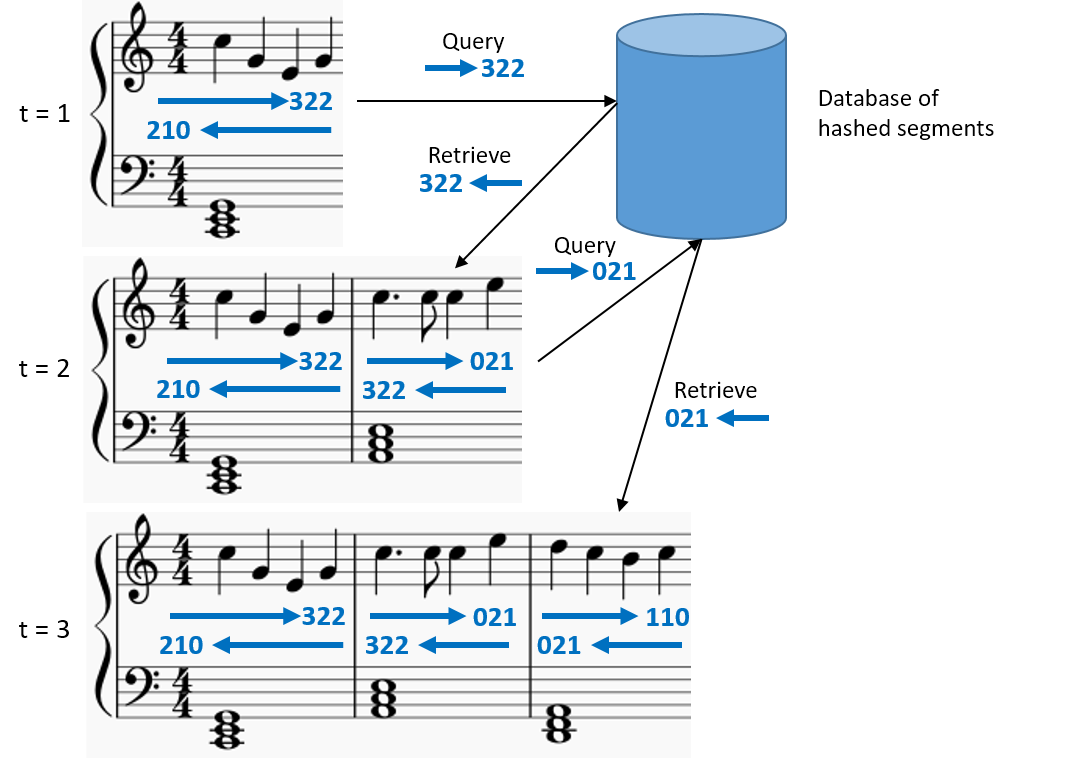}
	\caption{An example of the generation algorithm during the first 3 timesteps. At each timestep, the \B hash code of the last segment is used to retrieve a segment from the database with a similar \A hash code.}
	\label{fig:ex_gen}
\end{figure}

Fig. \ref{fig:ex_gen} shows the generation algorithm during the first three timesteps. Starting at $t = 1$, we have only the first segment in our new piece, $s_1$, which can be chosen randomly or from a pool of good ``starting" segments. Then, we use the \B hash code of $s_1$ (322 in the figure) to retrieve candidate segments with a matching (or similar) \A hash code. This process then repeats for a desired number of timesteps. Note that in practice, we use hash codes that are longer than 3 bits.

In summary, the goal of our model is to learn \B and \A hash functions that capture the \emph{composability} of any two segments $\si$ and $\sj$ without prior knowledge of music theory. The proposed network architecture is depicted in Fig. \ref{fig:network}. The model consists of two parallel networks, which aim to minimize/maximize the hamming distance between \hbi~ and \haj~ for \emph{composable}/\emph{non-composable} pairs. In the following subsections, we explain each component of the network, followed by the objective function and the pair-wise training method.

\subsection{LSTM Sequence Modeling} \label{sec:lstm}

Time-series data such as music can be analyzed by utilizing Recurrent Neural Networks (RNN) \cite{rnn}, which model the dynamics of an input sequence of timesteps through a sequence of hidden states, thereby learning the spatio-temporal structure of the input sequence. However, due to the exponential decay in retaining the content from each timestep in the input sequence, classical RNN's are limited in learning the long-term representation of sequences. Long Short-Term Memory (LSTM) \cite{lstm} was designed to overcome this limitation by finding the long-range dependencies between input sequences and the desired outputs. It accomplishes this by incorporating a sequence of memory cells, which are controlled by several types of gates that dictate what information enters and leaves the memory cells over time. Similar to \cite{lstm02,tiedparallel,pi,jambot,crnngan}, we opt to use LSTM networks to model the input time-series data, which in our work consists of an input segment of music. Note that the idea proposed in this paper does not restrict either the query or the retrieval segments to any specific length; thus, the specific choice to use the LSTM family of networks becomes more important as the segment length gets longer.

\begin{figure}
	\centering
	\includegraphics[height=6.5cm]{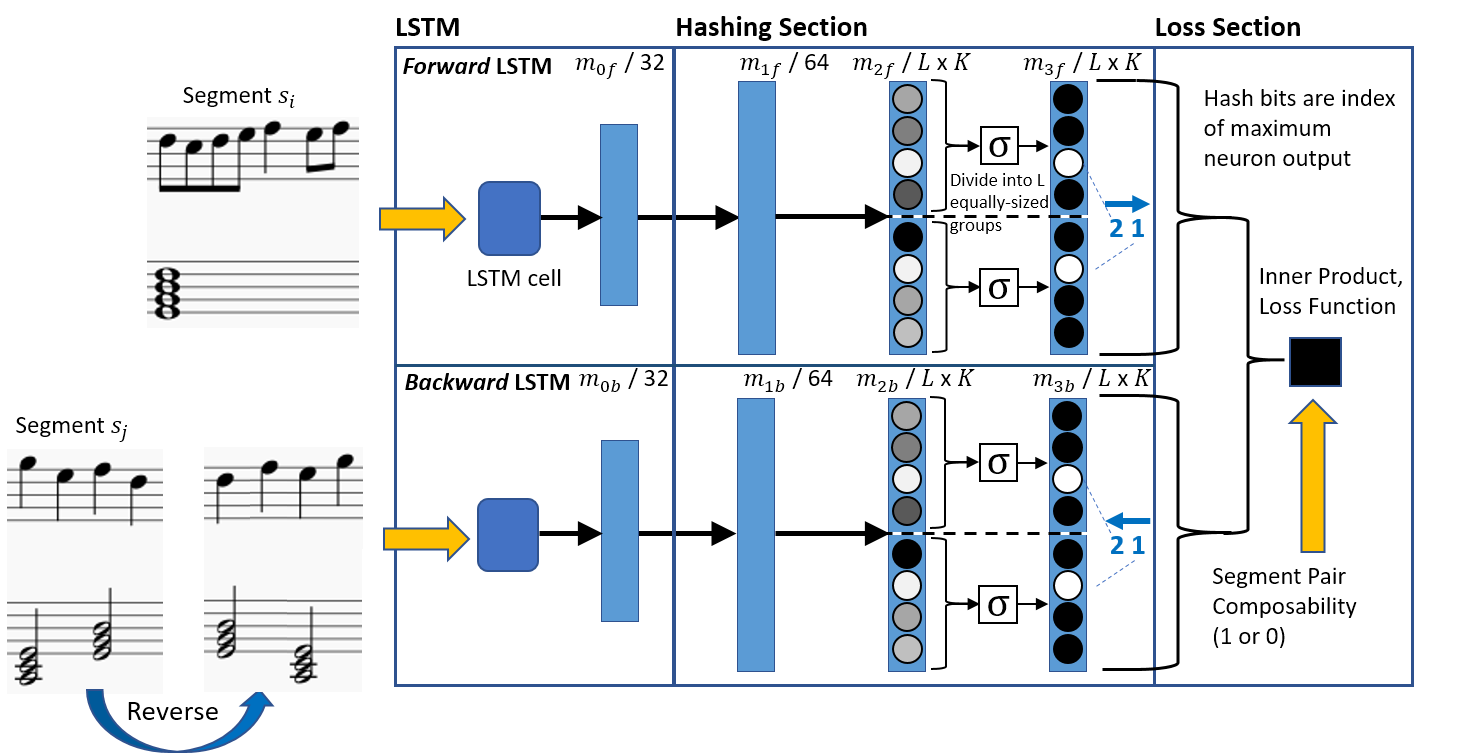}
	\caption{Network structure for deep segment hashing. Each layer is labeled above with a name (e.g. $m_{0f}$) and a dimension. The softmax function is given by $\sigma$ and is applied to groups of neurons of size $K$. Neuron values are depicted in grayscale, with darker colors indicating low values and lighter colors indicating high values (relative to the group of $K$ neurons). Assuming that the first neuron in each group is at index 0, both the \B hash code of $s_i$ and the \A hash code of $s_j$ would be the quaternary code $ 21 $.}
	\label{fig:network}
\end{figure}

From Fig. \ref{fig:network}, note that we use separate LSTM cells for the \B and \A networks. Given a segment pair ($\si$,$\sj$), the \B LSTM receives $\si$ fed sequentially in the forward direction while the \A LSTM receives the retrograde of $\sj$. Since LSTM's have the ability to memorize the content of an entire sequence, the last hidden state of the LSTM's ought to contain all the necessary information to represent each segment for the purpose of determining segment pair composability. Thus the last hidden state, denoted as $m_{0f}$ or $m_{0b}$ in Fig. \ref{fig:network}, becomes the input to the hashing section of the network.



\subsection{Ranking-Based Hashing} \label{sec:rankhash}


In this work, we adopt a category of hash functions known as ranking-based hashing. This class of hash functions outputs a $K$-ary hash code (as opposed to binary), meaning that each ``bit" chooses from one of $K$ different values. Ranking-based hashing has been used recently in multimedia hashing \cite{lsrh,ddsrh} due to the robustness of the technique against noise and variations \cite{wta}. The ranking-based hashing scheme finds a $K$-dimensional embedding of the input, which can be a linear or non-linear embedding, or a subspace of the input's original feature space. Then, one hash bit is set equal to the index of the maximum dimension of the embedding.


Formally, the ranking-based hash function, as described in \cite{lsrh,ddsrh} is defined as follows:

\begin{equation}
\label{eq:ranking}
\begin{gathered}
h(\mathbf z; \mathbf W) = \argmax\limits_{\mathbf h}\mathbf h^T \phi(\mathbf z; \mathbf W), \\
\text{s.t.} ~~ \mathbf h \in \{0, 1\}^K, \mathbf 1^T \mathbf h = 1,
\end{gathered}
\end{equation}


\noindent where $\mathbf z$ is the input example, $\mathbf W$ represents the learned parameters of the model, and $\phi(\mathbf z)$ is a $K$-dimensional embedding of the input. In our work, this nonlinear embedding is determined by the output of the neural network. Note that the output of the hash function is the 1-of-$K$ binary indicator vector $\mathbf h$, which represents the index of the maximum dimension of the data embedding $\phi$. Thus, the hash function $h$ in Eq. (\ref{eq:ranking}) encodes one $K$-ary hash ``bit". To obtain a hash code of length $L$, we need $L$ such hash functions.


In our work, the goal is to determine $K$-ary \B and \A hash codes that preserve the composability of the two segments. Given the last hidden state of the LSTM, we propose a hashing network similar to \cite{ddsrh} that is capable of learning $L$ independent $K$-dimensional embeddings of the input. In Fig. \ref{fig:network}, see that layer $m_{0*}$ (the LSTM hidden state) is fully-connected to the first layer of the hashing section, $m_{1*}$, of size 64. Layer $m_{1*}$ is then fully-connected to a second layer, $m_{2*}$, of size $L$ x $K$. This layer is then divided into $L$ groups of $K$ neurons; each group will represent one of the $K$-dimensional learned embeddings. Note that we apply the ReLU function after layers $m_{0*}$, $m_{1*}$, and $m_{2*}$.

Finally, we apply the softmax function to each of the $L$ groups in $m_{2*}$, resulting in layer $m_{3*}$. To understand why we use the softmax function, see that the hash function in Eq. (\ref{eq:ranking}) is discontinuous and non-convex due to the use of the argmax term. The softmax function serves as an approximation to the argmax term and allows us to reformulate Eq. (\ref{eq:ranking}) as:

\begin{equation}
\label{eq:reform}
h(\mathbf z; \mathbf W) \approx \sigma(\phi(\mathbf z; \mathbf W)),
\end{equation}

\noindent where $\sigma$ represents the softmax function. Note that the output of $h$ is still a $K$-dimensional vector, but it is now a continuous approximation of the discrete one-hot vector. This continuous version of the hash function makes it possible to optimize the loss function described in the following subsection.

\subsection{Objective Function} \label{sec:objfunc}

The objective function consists of a segment pair composability term and a bit balancing term. Each term is explained below, followed by the overall objective function.

\subsubsection{Segment Pair Composability}

For each pair of segments ($\si$, $\sj$), we define a binary \emph{composability} label $c_{ij}$ based on whether $\sj$ is suitable to follow $\si$. In this work, $c_{ij}$ is set to 1 if $\sj$ occurs after $\si$ in the dataset, and 0 otherwise. We then define the loss for a single training pair as

\begin{equation}
\label{eq:singlePairLoss}
l(\si,\sj) = \big( \frac{1}{L}\biB^T \bjA - c_{ij} \big)^2,
\end{equation}

\noindent where $\biB$, $\bjA \in [0,1]^{LK} = [h^1(\cdot)^T ~ h^2(\cdot)^T ~ .. ~ h^L(\cdot)^T]^T $ are vectorized continuous representations of the \B and \A hash codes for $\si$ and $\sj$ respectively. Here, $h$ is the hash function from Eq. (\ref{eq:reform}), which outputs a column vector whose elements sum to 1. $h^l(\cdot)$ denotes $h^l(\si;\WB)$ for $\si$ and $h^l(\sj;\WA)$ for $\sj$, and we use $\WB$ and $\WA$ to distinguish between the learned parameters for the \B and \A networks.


If $\si$ and $\sj$ are composable (that is, $c_{ij} = 1$), we would like the result of $\biB^T \bjA$ to be $L$, so that the result of the subtraction operation in Eq. (\ref{eq:singlePairLoss}) is 0. This will only happen if the positions of the '1' values in $\biB$ and $\bjA$ are the same, such that the indices of the maximum data embeddings agree for each hash bit. (Remember that since we are using the softmax function, these values will only approach '0' or '1' but are not discrete.) If $\si$ and $\sj$ are not composable ($c_{ij} = 0$), we would like the result of $\biB^T \bjA$ to be 0, so that again the result of the subtraction operation is 0. Together, these goals serve to minimize the hamming distance between the \B and \A codes for composable segment pairs, and maximize the same hamming distance for non-composable pairs.

The overall segment pair composability term is given as

\begin{equation}
\label{eq:simConst}
\mathcal{C}(\WB, \WA) = \big|\big| \frac{1}{L} \BB^T \BA - \mathbf{C} \big|\big|_E^2,
\end{equation}

\noindent where $\mathbf C$ is the binary composability matrix for all segment pairs, and $\BB$ and $\BA$ are the matrices of the binary vectorized \B and \A hash codes for all segments. $||*||_E$ represents the Euclidean norm (also known as the Frobenius norm).

\subsubsection{Bit Balancing}

We also incorporate a bit balancing term, which serves to balance the distribution of the observed values for each $K$-ary bit. For example, if bit 0 in every hash code takes the value `2', this bit does not contain useful or discriminative information. The details of the bit balancing term for $K$-ary hash codes are given in \cite{ddsrh}; here we omit the mathematical formulas and simply refer to the bit balancing term as $\mathcal B$. 

\subsubsection{Overall Objective Function}

The overall objective is thus to learn parameters $\WA$ and $\WB$ for the \B and \A networks respectively that minimize

\begin{equation}
\label{eq:obj}
\begin{gathered}
\min ~~ \mathcal{C}(\WB, \WA) + \alpha\mathcal{B}(\WB, \WA),
\end{gathered}
\end{equation}

\noindent where $\mathcal C$ is the composability term defined in Eq. (\ref{eq:simConst}), $\mathcal B$ is the bit balancing term, and $\alpha$ is a hyperparameter of the algorithm.

\subsection{Training the Network} \label{sec:training}

The network is trained in two stages: prediction pre-training and hash function learning. First, we train the \B and \A LSTM's to be able to accurately predict the next and previous timestep, respectively. For this task, we establish a cross entropy loss criterion for each of the one-hot segment features defined in section \ref{sec:featrep} (melody octave, melody pitch class, and chord class), as well as a binary cross entropy criterion for melody articulation. 

In the hash function learning stage, we initialize the LSTM weights with the values pre-trained on the prediction task and then allow them to be updated while the weights for the hashing component of the networks are learned. In this stage, we give segment pairs to the network along with their composability label (1 or 0). To the best of our knowledge, this method of pair-wise training has not been explored for music generation research. To adapt this task for the purpose of pair-wise segment composability, we define that \emph{positive} ($\si$,$\sj$) pairs are those that are composable, and \emph{negative} pairs are those that are not composable. We explain how these training pairs are chosen in section \ref{sec:pairs}.



\section{Experiments} \label{sec:experiments}

In this section, we discuss the experiments and interpret the results. First, we introduce the dataset in section \ref{sec:dataset}. We discuss how we represent the music sequences in section \ref{sec:featrep}. In section \ref{sec:pairs}, we explain how we choose positive and negative segment pairs for training the model. In section \ref{sec:trainingres}, we show how the hamming distance between the \B and \A hash codes evolves over the course of training for both positive pairs and negative pairs. Finally, in section \ref{sec:generations} we demonstrate the effectiveness of the hashing technique by showing two very qualitatively different pieces generated by the model. 


\subsection{Dataset} \label{sec:dataset}

In this work, we use a cleaned version of the Nottingham dataset\footnotemark. The dataset consists of over 1,000 folk tunes consisting of a melody and a chord in the bass. For simplicity, we only consider songs that are in 2/4 or 4/4 time and are in a major key, resulting in 460 songs. As in \cite{unitsel}, which found the ideal segment length to be one or two measures, we use 4-beat segments. We perform the 4-beat segmentation with a two beat overlap, meaning that the length of each segment is 4 beats, but each segment shares its first two beats with the preceding segment and the next two beats with the following segment. This results in a total of 11,608 segments in our dataset. We transpose all songs to C Major before performing the segmentation.
\footnotetext{Cleaned version is available for download at \url{https://github.com/jukedeck/nottingham-dataset}}

\subsection{Feature and Music Sequence Representation} \label{sec:featrep}

In accordance with the dataset used in this work, we assume that a song consists of a melody and a chord in the bass, similar to what one would see on a lead sheet of music. Since many compositions are written in lead sheet notation, we find this to be a good starting point for polyphonic music generation. After transposing all songs to C Major (and ignoring the minor mode), we identified 12 chords that occur frequently in the dataset. Thus, rather than using a higher-dimension piano roll representation, we opt for the low-dimensional representation described here, where we define 12 chord ``classes." Because the smallest subdivision of the beat is the sixteenth note, each beat is divided into four timesteps. (For simplicity we currently choose to ignore segments with triplet subdivisions.) Since segments consist of four beats, each segment has a length of 16 timesteps. Each timestep consists of a binary feature vector with 29 elements, summarized below:

\begin{itemize}
	\item 0: \textbf{Melody articulation}: 1 if the note in the melody is articulated at this timestep and 0 otherwise.
	\item 1-4: \textbf{Melody octave}: One-hot representation of the octave in which the melody is playing (all 0 if the melody is resting).
	\item 5-16: \textbf{Melody pitch class}: One-hot representation of the pitch class (C, C\#, D, etc.) of the melody note (all 0 if rest).
	\item 17-28: \textbf{Chord class}: One-hot representation of the chord class (all 0 if no chord).

\end{itemize}

We omit a ``chord articulation" bit because the vast majority of chords occur on beat 1 or 3 of the segment and are sustained for increments of 2 beats.


\subsection{Positive and Negative Pairs} \label{sec:pairs}

The number of positive training pairs is approximately equal to the number of segments in the dataset (11,608), since each positive pair consists of a segment and the segment immediately following it in the same song. Given this definition, each segment only exists in two positive pairs. Thus, the number negative pairs is approximately equal to the number of segments in the dataset squared: about 121 million pairs. Unfortunately, there is no strict definition for what comprises a ``positive" pair or a ``negative" pair. In fact, many of the so-called ``negative" pairs should in fact be classified as positive pairs and would only serve to confuse the network during training.

In this work, we adopt a simple statistical method to determine how likely it is that a ``negative" pair is in fact a ``positive" pair. For a segment pair ($\si$,$\sj$), let $u_i$ and $v_i$ be the last chord and last melodic pitch (C4, for example) in segment $\si$, respectively, and $u_j$ and let $v_j$ be the first chord and first melodic pitch in segment $\sj$, respectively. We define $P(u_j | u_i)$ as the probability of observing $u_j$ given $u_i$, and we similarly define $P(v_j | v_i)$ for melodic pitches. Now, by assuming that all ``negative" pairs are in fact negative, we analyze positive and negative pairs separately in order to calculate $P(u_j | u_i, +)$ and $P(u_j | u_i, -)$. We then define the probability that a pair is a positive pair as

\begin{equation}
P(+ | u_i, u_j, v_i, v_j) = [P(+|u_i,u_j) + P(+|v_i,v_j)] / 2,
\label{eq:probpos}
\end{equation}

\noindent where

\begin{equation}
P(+|u_i,u_j) = \frac{P(u_j|u_i,+)}{P(u_j|u_i,+) + P(u_j|u_i,-)},
\end{equation}

\begin{equation}
P(+|v_i,v_j) = \frac{P(v_j|v_i,+)}{P(v_j|v_i,+) + P(v_j|v_i,-)}.
\end{equation}

Given Eq. (\ref{eq:probpos}), we can set a threshold for the ``negative" pairs in order to filter out those pairs that a human might label as a ``positive" pair. In our experiments we set the threshold to 0.5. Thus, any ``negative" pair with $P(+ | u_i, u_j, v_i, v_j) < 0.5$ is used as a negative pair during training, and any ``negative" pair with $P(+ | u_i, u_j, v_i, v_j) >= 0.5$ is removed from training to avoid confusing the network.

\subsection{Training Results} \label{sec:trainingres}

\begin{figure}[]
	\centering
	\subfloat[Training Pairs]{
		\includegraphics[height=4.75cm, width=5.3cm]{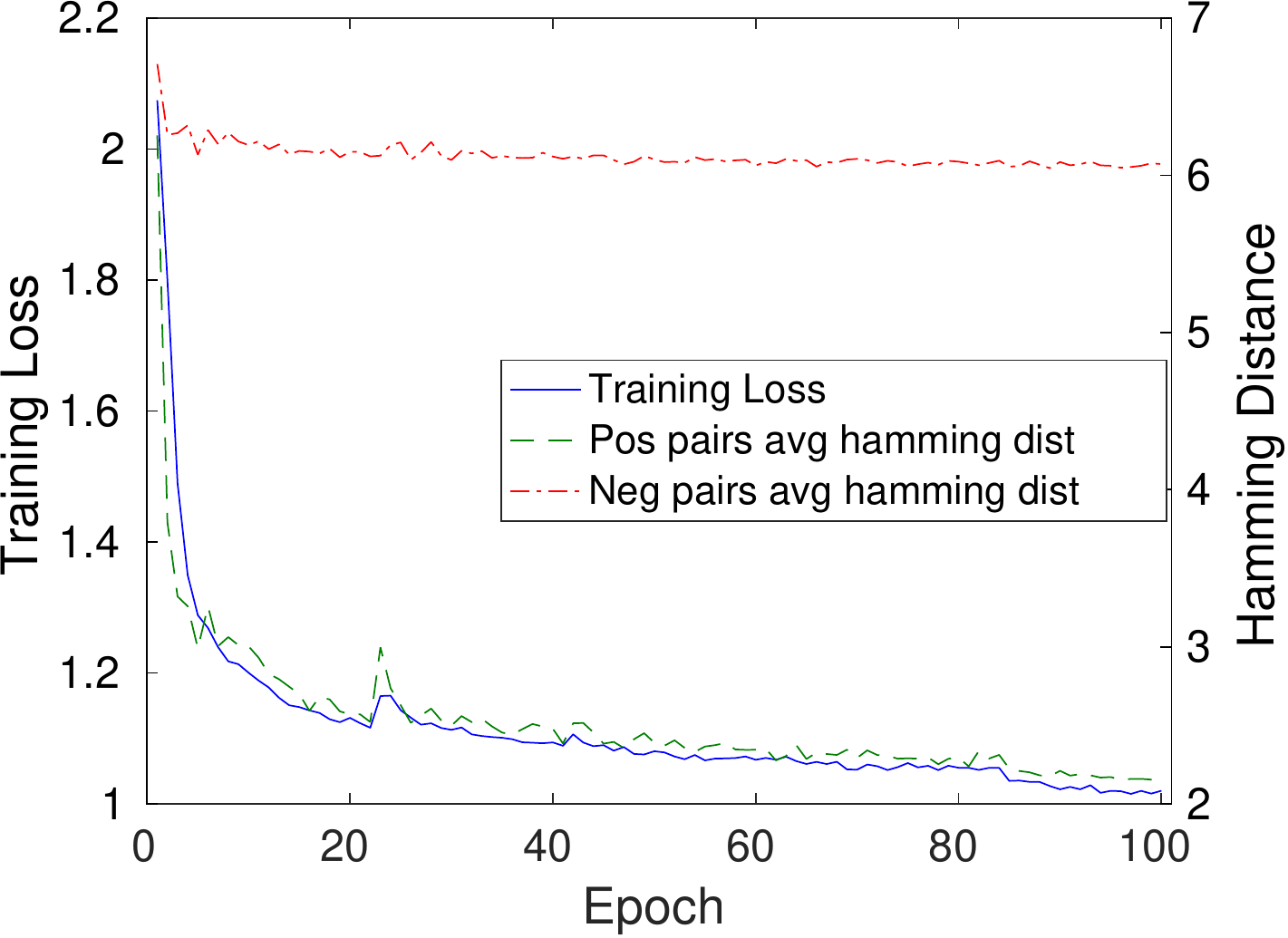}
	}
	\qquad
	\subfloat[Validation Pairs]{
		\includegraphics[height=4.75cm, width=5.3cm]{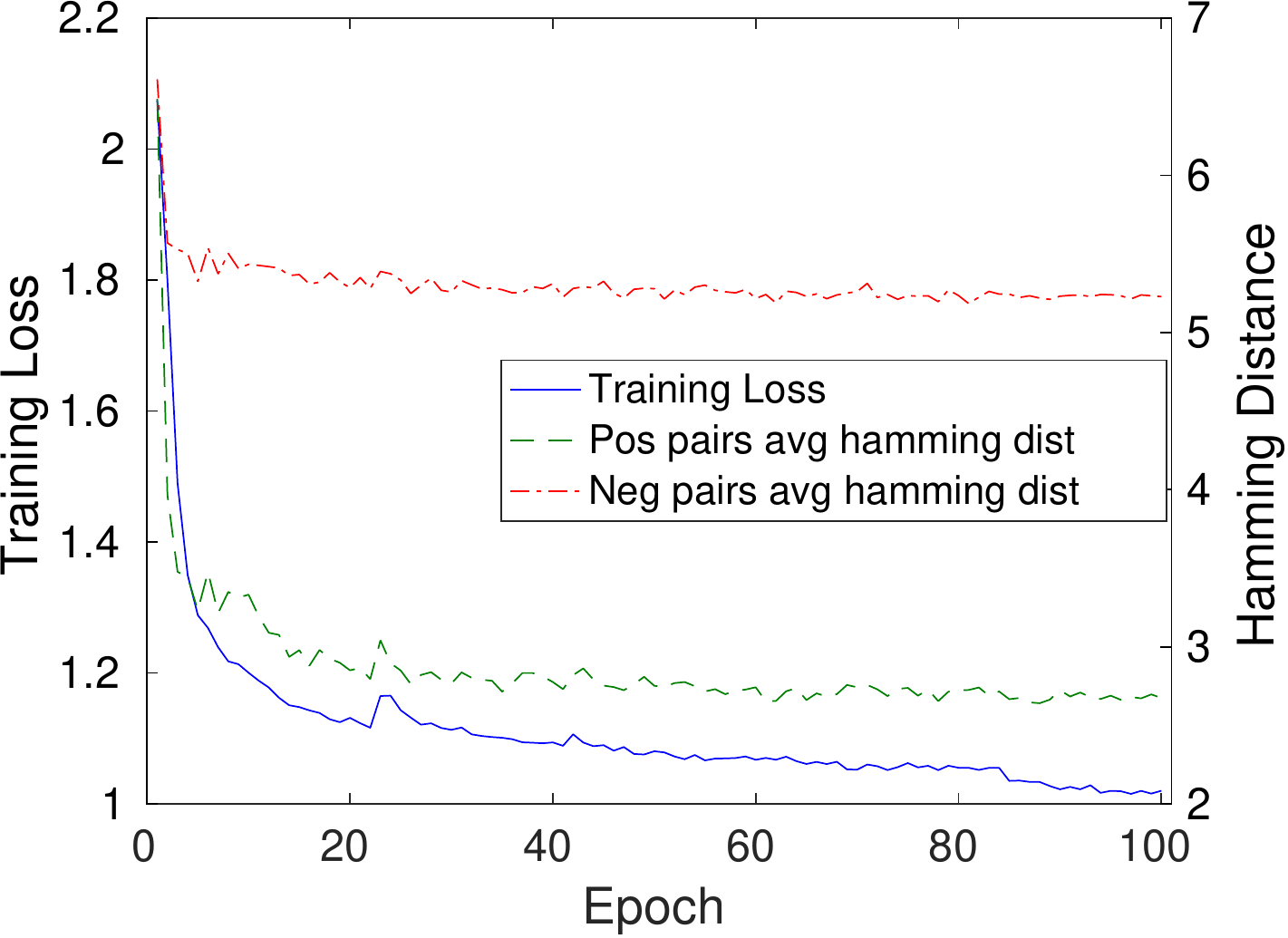}
	}
	\caption{Average hamming distance between \hbi~ and \haj~ for positive and negative \emph{training} pairs (a) and \emph{validation} pairs (b), observed while training for 100 epochs. In this experiment, $L = 8$ and $K = 4$. Training loss is also shown.}
	\label{fig:training_pairs}
\end{figure}

In this section, we analyze the average hamming distance between \hbi~ and \haj~ for all pairs ($\si$,$\sj$) observed while training the network for 100 epochs. Here, we use an 8-bit hash code ($L = 8$), where each hash bit can take 4 values ($K = 4$). Note that hamming distance works the same way for $K$-ary codes as for binary codes: given two hash codes, the hamming distance is the number of bit positions that have different values. Fig. \ref{fig:training_pairs} shows the results of this experiment, plotted alongside the training loss. Fig. \ref{fig:training_pairs}(a) plots the average hamming distance between \hbi~ and \haj~ for all positive and negative \emph{training} pairs. We withhold 750 positive pairs and 750 negative pairs from training to form a validation set that tests the generalizability of the model to segment pairs that it has not seen. This generalizability is an important attribute of the model, because it allows the creation of new music that still conforms to the implicit rules learned during training. Fig. \ref{fig:training_pairs}(b) plots the average hamming distance between \hbi~ and \haj~ for all positive and negative \emph{validation} pairs.

In Fig. \ref{fig:training_pairs}(a), we see that the average hamming distance between \hbi~ and \haj~ for the positive training pairs decreases over time, which is desired. Although the average hamming distance between \hbi~ and \haj~ for negative training pairs does not appear to increase over time, it remains relatively high throughout the course of training. Because the hamming distance for the negative pairs remains fairly constant, the contribution of the negative pairs to the training loss is small. Thus, the training loss closely mirrors the hamming distance for the positive training pairs. In Fig. \ref{fig:training_pairs}(b), we see similar results for the positive and negative \emph{validation} pairs, although the gap in average hamming distance is not as large as for the training pairs.


\subsection{Example Generations} \label{sec:generations}

\begin{figure}
	\centering
	\subfloat[\textbf{Task 1}: composition created by retrieving segments with \textbf{minimum} hamming distance between \B and \A hash codes (normal use case).]{
		\includegraphics[height=5cm, width=11cm]{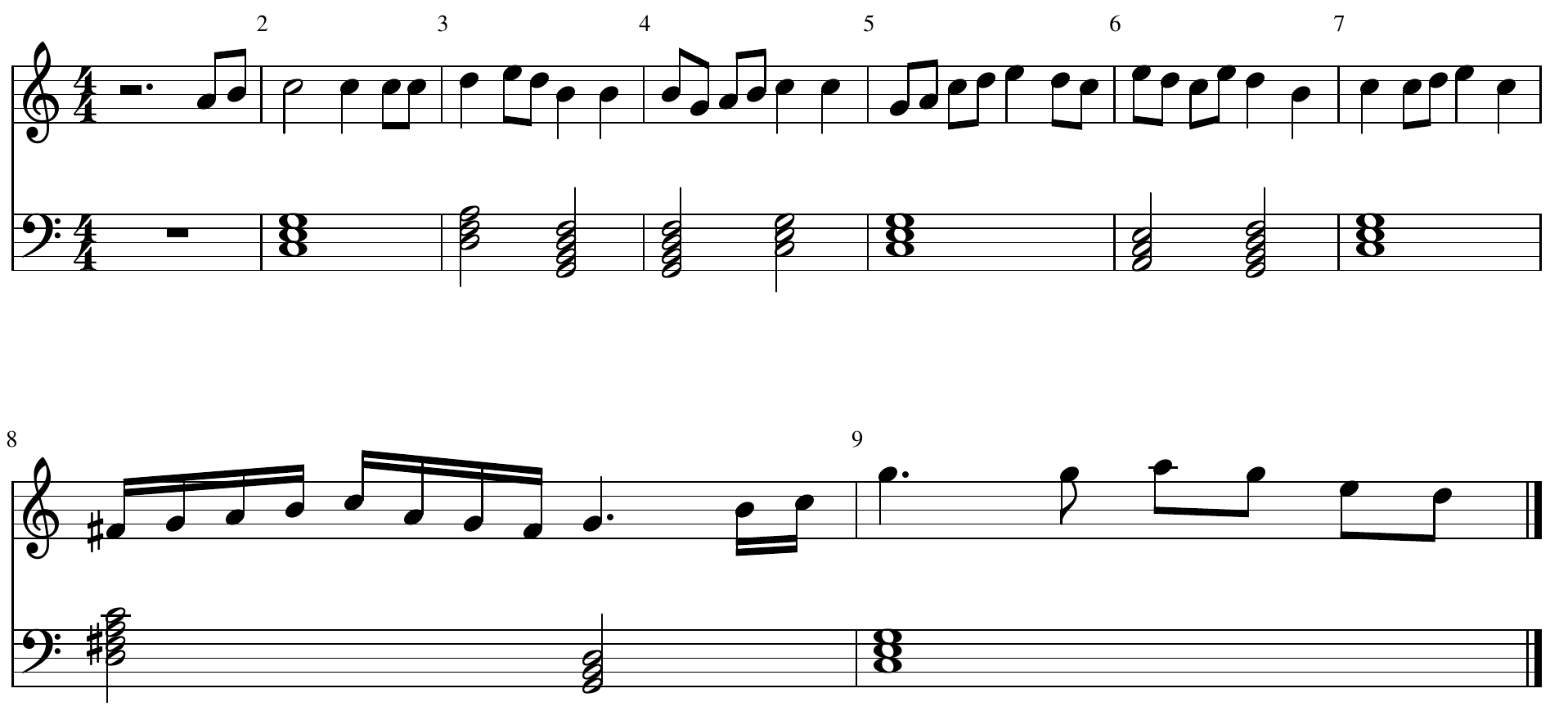}
	}

	\subfloat[\textbf{Task 2}: composition created by retrieving segments with \textbf{maximum} hamming distance between \B and \A hash codes.]{
		\includegraphics[height=5cm, width=11cm]{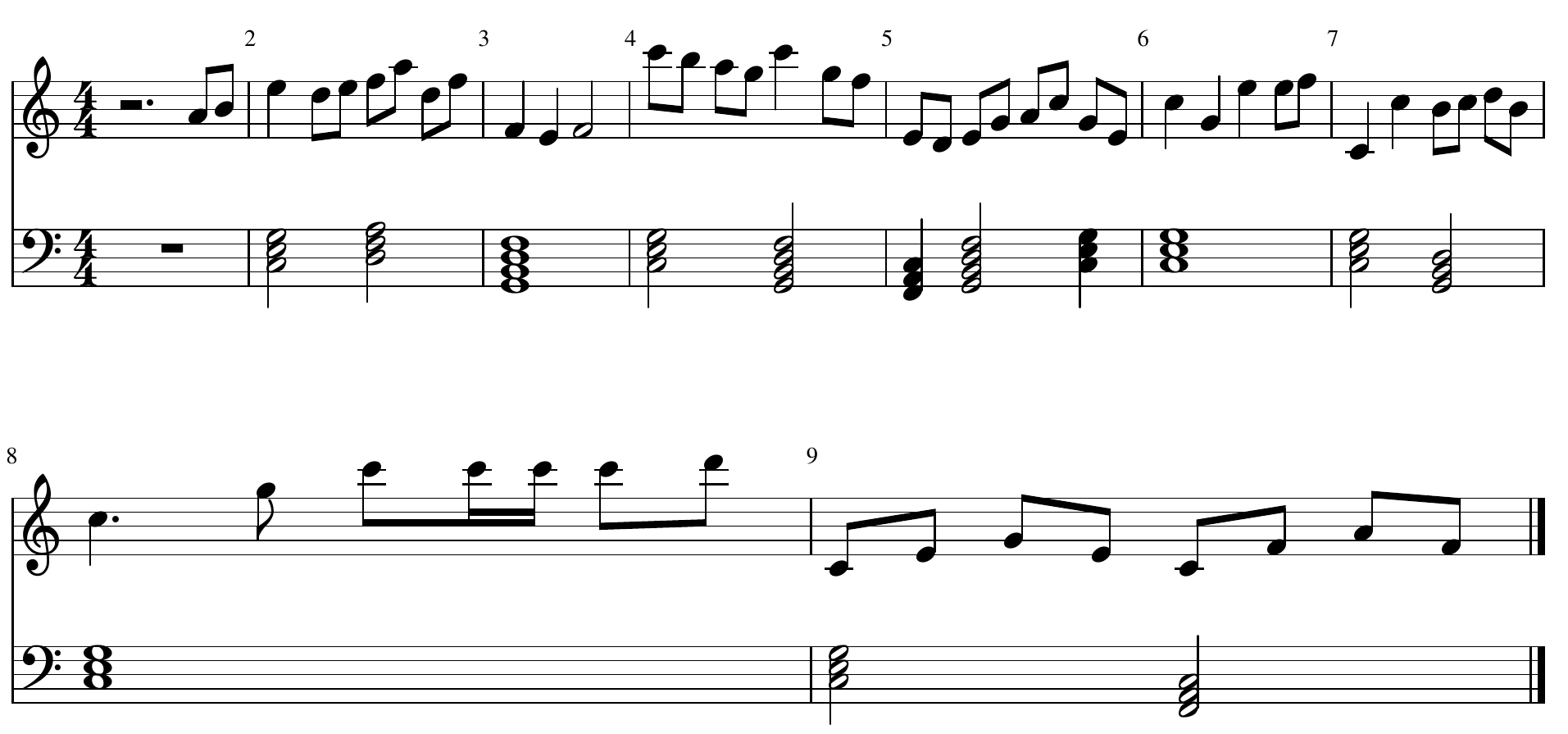}
	}
	\caption{Example compositions created by minimizing (a) and maximizing (b) the hamming distance between the \B and \A hash codes. (a) is coherent and flows smoothly, while (b) leaps around too much in the melody and does not flow well.}
	\label{fig:gens}
\end{figure}

In Fig. \ref{fig:gens}, we show two short compositions that were created using our algorithm. First, we show an example that was created using the aforementioned procedure: we query the database by using the \B hash code of the most recently chosen segment to retrieve candidate segments with \A hash codes that are of the minimum hamming distance away from the \B code. If multiple candidate segments are found, we choose one at random. Next, we use the same starting segment and show an example that was created by performing the opposite retrieval task: retrieving segments with \A hash codes that are of the \emph{maximum} hamming distance away from the query \B code. Note that these examples were generated by retrieving 8 continuation segments, one at a time, and we currently do not place any requirements on the finality of the last segment. However, we do place a restriction on the number of segments that can come from the same song in the dataset. This is intended to ensure the originality of the generated music so that the algorithm does not copy too much from any given song.

From Fig. \ref{fig:gens}(a), we see that using the algorithm in the intended fashion (\emph{minimizing} hamming distances between adjacent segments) results in music that flows well and sounds pleasing. On the other hand, by \emph{maximizing} the hamming distance between the segments as in Fig. \ref{fig:gens}(b), we obtain music that does not flow well, especially in the melody. These results agree with the objective of DSHL, which is to minimize the hamming distance between the \B and \A hash codes of composable segments, and to maximize the hamming distance between these hash codes for non-composable segments. Of course, not every transition demonstrates the fulfillment of this objective. For example, in Fig. \ref{fig:gens}(b), we expect each segment-to-segment transition to be a poor transition. However, measure 8 could plausibly follow measure 7, which represents a failure case.

\section{Conclusion and Future Work} \label{sec:conclusions}

In this paper, we have proposed Deep Segment Hash Learning (DSHL), a new model for music generation that uses LSTM's and ranking-based hashing to learn a compact representation for musical segments that captures the composability between pairs of segments. By assigning \B and \A hash codes to each segment in database, we can achieve fast retrieval of candidate segments to continue a piece of music that is being created. We have demonstrated that DSHL is an effective means to generate music that is both original and pleasing to the ear. To the best of our knowledge, our method is the first to use segment hash learning for music generation. Additionally, DSHL is the first method to use pair-wise training on segments of music. Thus, this work sheds light on a new and promising way to generate music using deep learning and segment hashing. 

Given the recent popularity of Generative Adversarial Networks (GAN) \cite{gan} and their application to music generation, we are inspired to incorporate GAN's into our deep segment hash learning framework and to introduce semi-supervised learning to broaden the pool of positive training pairs available for training. Finally, we plan to experiment by using segments of different lengths and additional methods to improve the global structure and coherence of the generated music.

\bibliographystyle{splncs}
\bibliography{musicbib}

\begin{thebibliography}{10}

\bibitem{cope}
Cope, D., Mayer, M.J.:
\newblock Experiments in musical intelligence. Volume~12.
\newblock AR editions Madison, WI (1996)

\bibitem{unitsel}
Bretan, M., Weinberg, G., Heck, L.:
\newblock A unit selection methodology for music generation using deep neural
  networks.
\newblock arXiv preprint arXiv:1612.03789 (2016)

\bibitem{improtek}
Nika, J., Chemillier, M., Assayag, G.:
\newblock Improtek: introducing scenarios into human-computer music
  improvisation.
\newblock Computers in Entertainment (CIE) \textbf{14} (2016) ~4

\bibitem{continuator}
Pachet, F.:
\newblock The continuator: Musical interaction with style.
\newblock Journal of New Music Research \textbf{32} (2003)  333--341

\bibitem{hashingcont}
Charapko, A., Chuan, C.H.:
\newblock Indexing and retrieving continuations in musical time series data
  using relational databases.
\newblock In: Multimedia (ISM), 2014 IEEE International Symposium on, IEEE
  (2014)  341--346

\bibitem{lsrh}
Li, K., Qi, G., Ye, J., Hua, K.:
\newblock Linear subspace ranking hashing for cross-modal retrieval.
\newblock IEEE transactions on pattern analysis and machine intelligence (2016)

\bibitem{dcmh}
Jiang, Q.Y., Li, W.J.:
\newblock Deep cross-modal hashing.
\newblock In: The IEEE Conference on Computer Vision and Pattern Recognition
  (CVPR). (2017)

\bibitem{ddsrh}
Joslyn, K., Li, K., Hua, K.A.:
\newblock Cross-modal retrieval using deep de-correlated subspace ranking
  hashing.
\newblock In: ACM International Conference on Multimedia Retrieval (ICMR).
  (2018)

\bibitem{rnn}
Elman, J.L.:
\newblock Finding structure in time.
\newblock Cognitive science \textbf{14} (1990)  179--211

\bibitem{lstm}
Hochreiter, S., Schmidhuber, J.:
\newblock Long short-term memory.
\newblock Neural computation \textbf{9} (1997)  1735--1780

\bibitem{lstm02}
Eck, D., Schmidhuber, J.:
\newblock Finding temporal structure in music: Blues improvisation with lstm
  recurrent networks.
\newblock In: Neural Networks for Signal Processing, 2002. Proceedings of the
  2002 12th IEEE Workshop on, IEEE (2002)  747--756

\bibitem{tiedparallel}
Johnson, D.D.:
\newblock Generating polyphonic music using tied parallel networks.
\newblock In: International Conference on Evolutionary and Biologically
  Inspired Music and Art, Springer (2017)  128--143

\bibitem{pi}
Chu, H., Urtasun, R., Fidler, S.:
\newblock Song from pi: A musically plausible network for pop music generation.
\newblock arXiv preprint arXiv:1611.03477 (2016)

\bibitem{jambot}
Brunner, G., Wang, Y., Wattenhofer, R., Wiesendanger, J.:
\newblock Jambot: Music theory aware chord based generation of polyphonic music
  with lstms.
\newblock arXiv preprint arXiv:1711.07682 (2017)

\bibitem{crnngan}
Mogren, O.:
\newblock C-rnn-gan: Continuous recurrent neural networks with adversarial
  training.
\newblock arXiv preprint arXiv:1611.09904 (2016)

\bibitem{dssm}
Gao, J., Deng, L., Gamon, M., He, X., Pantel, P.:
\newblock Modeling interestingness with deep neural networks (2014) US Patent
  App. 14/304,863.

\bibitem{lsh}
Indyk, P., Motwani, R.:
\newblock Approximate nearest neighbors: towards removing the curse of
  dimensionality.
\newblock In: Proceedings of the thirtieth annual ACM symposium on Theory of
  computing, ACM (1998)  604--613

\bibitem{boulanger}
Boulanger-Lewandowski, N., Bengio, Y., Vincent, P.:
\newblock Modeling temporal dependencies in high-dimensional sequences:
  Application to polyphonic music generation and transcription.
\newblock arXiv preprint arXiv:1206.6392 (2012)

\bibitem{rnnrbm}
Boulanger-Lewandowski, N., Bengio, Y., Vincent, P.:
\newblock Modeling temporal dependencies in high-dimensional sequences:
  Application to polyphonic music generation and transcription.
\newblock arXiv preprint arXiv:1206.6392 (2012)

\bibitem{algornn}
Colombo, F., Muscinelli, S.P., Seeholzer, A., Brea, J., Gerstner, W.:
\newblock Algorithmic composition of melodies with deep recurrent neural
  networks.
\newblock arXiv preprint arXiv:1606.07251 (2016)

\bibitem{waite}
Waite, E.:
\newblock Generating long-term structure in songs and stories.
\newblock Magenta Bolg (2016)

\bibitem{brnn}
Berglund, M., Raiko, T., Honkala, M., K{\"a}rkk{\"a}inen, L., Vetek, A.,
  Karhunen, J.T.:
\newblock Bidirectional recurrent neural networks as generative models.
\newblock In: Advances in Neural Information Processing Systems. (2015)
  856--864

\bibitem{concert}
Mozer, M.C.:
\newblock Neural network music composition by prediction: Exploring the
  benefits of psychoacoustic constraints and multi-scale processing.
\newblock Connection Science \textbf{6} (1994)  247--280

\bibitem{deepbach}
Hadjeres, G., Pachet, F., Nielsen, F.:
\newblock Deepbach: a steerable model for bach chorales generation.
\newblock arXiv preprint arXiv:1612.01010 (2016)

\bibitem{wta}
Yagnik, J., Strelow, D., Ross, D.A., Lin, R.s.:
\newblock The power of comparative reasoning.
\newblock In: Computer Vision (ICCV), 2011 IEEE International Conference on,
  IEEE (2011)  2431--2438

\bibitem{gan}
Goodfellow, I., Pouget-Abadie, J., Mirza, M., Xu, B., Warde-Farley, D., Ozair,
  S., Courville, A., Bengio, Y.:
\newblock Generative adversarial nets.
\newblock In: Advances in neural information processing systems. (2014)
  2672--2680

\end{thebibliography}

\end{document}